\documentclass[10pt,conference]{IEEEtran}
\usepackage{filecontents}

\usepackage{adjustbox}
\usepackage{multirow}
\usepackage{subfigure}
\usepackage{color}

\usepackage{cite}
\usepackage[cmex10]{amsmath}
\usepackage{amsthm}

\usepackage{algorithmic}
\usepackage{array}
\ifCLASSOPTIONcompsoc
  \usepackage[caption=false,font=normalsize,labelfont=sf,textfont=sf]{subfig}
\else
  \usepackage[caption=false,font=footnotesize]{subfig}
\fi
\usepackage{url}
\begin{document}
\title{Using Big Five Personality Model to Detect Cultural Aspects in Crowds}

%------------------------------------------------------------------------- 
% change the % on next lines to produce the final camera-ready version 
\newif\iffinal
%\finalfalse
\finaltrue
\newcommand{\jemsid}{72}
%------------------------------------------------------------------------- 


\iffinal

\author{\IEEEauthorblockN{Rodolfo Favaretto, Leandro Dihl, Soraia Raupp Musse}
\IEEEauthorblockA{Pontifical Catholic University of Rio Grande do Sul\\
Graduate Studies on Computer Science\\
Porto Alegre, RS, Brazil\\
Email: soraia.musse@pucrs.br}
\and

\IEEEauthorblockN{Felipe Vilanova and Angelo Brandelli Costa}
\IEEEauthorblockA{Pontifical Catholic University of Rio Grande do Sul\\
Graduate Studies on Psychology\\
Porto Alegre, RS, Brazil\\
Email: angelo.costa@pucrs.br}
}


%Telephone: (800) 555--1212, Fax: (888) 555--1212}

\else
  \author{Sibgrapi paper ID: \jemsid \\ }
\fi

\maketitle

\begin{abstract}
The use of information technology in the study of human behavior is a subject of great scientific interest. Cultural and personality aspects are factors that influence how people interact with one another in a crowd. This paper presents a methodology to detect cultural characteristics of crowds in video sequences. Based on filmed sequences,  pedestrians are detected, tracked and characterized. Such information is then used to find out cultural differences in those videos, based on the Big-five personality model. Regarding cultural differences of each country, results indicate that this model generates coherent information when compared to data provided in literature.
\end{abstract}


\IEEEpeerreviewmaketitle

\section{Introduction}
\label{sec:intro}

Crowd analysis is a phenomenon of great interest in a large number of applications. Surveillance, entertainment and social sciences are examples of fields that can benefit from the development of this area of study. Literature dealt with different applications of crowd analysis, for example counting people in crowds~\cite{Chan2009, cai2014}, group and crowd movement and formation~\cite{Solmaz2012, Zhou2014, Ricky:15, jo2013review} and detection of social groups in crowds~\cite{solera_2013, Shao2014, Feng2015, Chandran2015}. Normally, these approaches are based on personal tracking or optical flow algorithms, and handle as features: speed, directions and distances over time. Recently, some studies investigated cultural difference in videos from different countries. Chattaraj et al.~\cite{CHATTARAJ2009} suggested that cultural and population differences could produce deviations in speed, density and flow of the crowd. Favaretto et al.~\cite{Favaretto:2016} discussed  cultural dimensions according to Hofstede analysis~\cite{Hofstede:2011} and presented a methodology to map data from video sequences to the dimensions of Hofstede cultural dimensions theory.

In this paper, we propose to detect crowd-cultural aspects based on the Big-five personality model (or OCEAN)~\cite{costa07} (Brazilian version) from NEO PI-R~\cite{costa92} using individuals behaviors automatically detected in video sequences. For this, we used the NEO PI-R~\cite{costa07} that is the standard questionnaire measure of the Big-Five Factor Model. The questionnaire provides a detailed personality description that can be a valuable resource for a variety of professionals. We firstly selected NEO PI-R items related to individual-level crowd characteristics and the corresponding factor, as described later in this paper. For example: "Like being part of crowd at sporting events" corresponding to the factor “Extroversion”. More details about personality models are discussed in Section~\ref{sec:related}.
%SO: Angelo vê essa coisa da versão Brasil que coloquei acima... OK!

After the NEO PI-R items selection (related to crowds characteristics), we propose a way to map data extracted from video sequences to Big-Five parameters, as described in Section~\ref{sec:model}. 

Since there are different distributions of each of the Big-Five factors in different countries~\cite{costa07}, we hypothesize that it would be possible to detect cultural differences  from videos processing crowd behavior from different countries. 
This discussion is addressed in Section~\ref{sec:results}. Conclusions and future work are presented in Section~\ref{sec:conclusions}.

\section{Related Work}
\label{sec:related}

This section discusses some topics concerned with personality and also associated with crowd simulation.

Personality may be labeled as deep psychological individual level trait~\cite{cattell50}. Trait is an inference made after observed behaviors that seeks to explain its regularity ~\cite{hall98}. In general, researchers agree that there are five robust orthogonal traits which effectively matched personality attributes~\cite{digman90}, known as the Big Five: Openness to experience (“the active seeking and appreciation of new experiences”); Conscientiousness (“degree of organization, persistence, control and motivation in goal directed behavior”); Extraversion (“quantity and intensity of energy directed outwards in the social world”); Agreeableness (“the kinds of interaction an individual prefers from compassion to tough mindedness”); Neuroticism (how much prone to psychological distress the individual is)~\cite{lordw07}. %(adaptar refereências) AQUI não entendi!
The development of the Big Five personality model has its roots in the work by Allport and Odbert (1936) who tried to identify individual differences extracting relevant words in the Webster’s Unabridged Dictionary. They worked with the hypothesis that the most important individual differences would be coded in language, since as they are the most important, there would be an evolutionary necessity to communicate it. Although Allport and Odbert (1936) found 4.500 words which referred to generalized and stable personality traits, their technique couldn’t originate few personality traits which explained most part of behaviors variance.

Raymond Cattel is commonly referred as the one who developed the methodology which permitted the objective grouping of hundreds of trait descriptors in a set of higher level factors~\cite{digman90}. Cattell~\cite{cattell48} developed a taxonomy of individual differences that consisted of 16 primary factors and 8 second-order factors. Nevertheless, attempts to replicate his work were unsuccessful~\cite{fiske48} and researchers agreed that only the 5-factor model matched his data, originating the Big Five personality model.

The NEO PI-R~\cite{costa92} is one of the most used instrument based on the Big Five personality theory. It assesses the normal adult personality and is internationally recognized as a gold standard for personality assessment. One of its advantages is that it further specifies six facets within each personality trait and have data from several countries which easily allows cross-cultural comparisons~\cite{McCrae2002,McCrae05}. Although the current empirical evidence matching individual level traits, such personality and crowd behavior is not strong (one of the few examples is~\cite{Barry97}), the Big-Five personality model is widely used to model computational crowd simulation~\cite{kaup06, Durupinar08, Guy11}. The model allows to simulate a crowd with individual level parameters based on the expected behaviors of the agents.

Recently, research has shown that digital records can be an effective tool in predicting personality traits. Facebook likes, for example, can predict the actual score of the Big-Five personality model, especially the Openness trait \cite{Kosinski:2013}, providing roughly as much information as the self-reported personality test score itself. This makes room for the use of computational methods in predicting an individual’s personality as effectively as through the analysis of self-reported scores. A computational method to assess personality score can be also useful since there are issues concerning traditional self-report techniques: 1) individuals may deceive themselves and unintentionally distort their ratings of socially desirable traits in a positive direction \cite{paulhus:1992}; 2) individuals can fake their responses to personality measures, especially in contexts which the test is used as a selection criterion, such as in job interviews; 3) individuals can distort their answers in different levels and ways, making it harder to apply a general statistical correction which serves equally to everyone \cite{oh:2011}.

One effective alternative to the self-report method is the observer ratings of personality (i.e., acquaintances, friends, colleagues). A meta-analysis has shown that observational rating provides substantial incremental validity over self-reports of personality \cite{oh:2011}. One of the possible reasons for it is that self-reports assess the internal dynamics of an individual, whereas observer ratings analyze the  behavioral performance. As the behavior is a better predictor of the future performance than the inner dynamics of an individual \cite{JASP1355} it might be the reason of the better predictive validity. Therefore, we propose that it is possible to predict facets of personality traits of individuals through computer vision of crowd behavior as effectively as through the self-report method and observer ratings such as a collegue or a friend. The rationale behind this proposal is that since observer ratings might be as valid as the self-report, computer vision might be effective as well - since the behavioral component is being analyzed and not solely the inner dynamics of an individual. One example is the way we can successfully predict players’ personality scores through behavioural cue of their avatars in virtual worlds and games~\cite{Yee:2011,Yee:2011-2}. %citar Yee, N., Ducheneaut, N., Nelson, L., & Likarish, P. (2011, May). Introverted elves & conscientious gnomes: the expression of personality in world of warcraft. In Proceedings of the SIGCHI Conference on Human Factors in Computing Systems (pp. 753-762). ACM. --- OK

% e citar também  Yee, N., Harris, H., Jabon, M., & Bailenson, J. N. (2011). The expression of personality in virtual worlds. Social Psychological and Personality Science, 2(1), 5-12. --- OK

Concerning cultural simulation, Lala et al.~\cite{Lala2012} introduced a virtual environment that allows the creation of different types of cultural crowds. The crowd parameterization is based on the cultural dimensions presented by Hofstede~\cite{Hofstede:1991}. The work proposed by Kaminka~\cite{Kaminka:2011} presents data that aim to differentiate populations with regard to their behavior of movement in crowds. Cultural parameters are proposed and analyzed in videos from different countries, for later comparison. Some of the analyzed parameters are: speed, personal space, collision quantities and population flow.

In this paper, the idea is to map parameters from individual behaviors (automatically detected from video sequences of different countries) to generate a Big-Five personality model score (OCEAN)~\cite{costa07} for each of them. In this sense, our contribution is a model based on a set of equations that handle the individual parameters related to crowd behaviors obtained from videos and mapped to crowd-related Big-Five personality traits, generating profiles of each individual/analyzed video. Since personality differences in the Big-five model between countries are established in the literature ~\cite{McCrae2002,McCrae05}, one can compare each specific result extracted from the video with the related country/cultural score.

\section{The Proposed Approach}
\label{sec:model}

Our model presents two main steps: video data extraction and cultural analysis. The first step aims to obtain the individual trajectories from observed pedestrians in real videos. Using these trajectories, we extracted data that are useful for the second step, that is responsible for the personality and cultural analysis. 

\subsection{Individuals Data Extraction}

Initially, the information about  people from real videos is obtained using a tracker~\cite{Bins2013} to recover people trajectories. The features are following described.  We compute firstly the geometric information for each person $i$ at each timestep: \textit{i)} 2D position $x_i$ (meters); \textit{ii)} speed $s_i$ (meters/frame); \textit{iii)} angular variation $\alpha_i$ (degrees) w.r.t. a reference vector $\vec{r}=(1,0)$. In addition, three other features are also computed: \textit{iv)} collectivity $\phi_i$, \textit{v)} socialization $\vartheta_i$ and \textit{vi)} isolation levels $\varphi_i$. These features were chosen because two reasons: Firstly, they are strongly related with the questions concerned with groups activities in Neo-Pi survey~\cite{costa07}. The second reason is the theory behind socialization/isolation that easily can be represented through geometric data (positions and distances), and collectivity that has been already explored in the context of crowd behaviors detection~\cite{Zhou2014}.

To compute the collectivity affecting one individual $i$ from all $n_i$ individuals in his/her social space (as presented in~\cite{Favaretto:Sib:2016}), we used Equation~\ref{eq:colec}:

\begin{equation}
\phi_i = \sum_{j=0}^{n-1} \gamma e^{(-\beta \varpi(i,j)^{2})}, 
\label{eq:colec}
\end{equation}

where the collectivity between two individuals $i$ and $j$ is calculated as a decay function of $\varpi(i,j) = s(s_i,s_j).w_1+o(\alpha_i,\alpha_j).w_2$, considering $s$ and $o$ respectively the speed and orientation differences between the two people  and $w_1$ and $w_2$ are constants that should regulate the offset in meters and radians. We have used $w_1=1$ and $w_2=1$. So, values for $\varpi(i,j)$ are included in interval $0\leq \varpi(i,j) \leq 4.34$. $\gamma = 1$ is the maximum collectivity value when $\varpi(i,j)=0$, and $\beta = 0.3$ is empirically defined as decay constant. Hence, $\phi_{i}$ is a value in the interval $[0;1]$.

\begin{figure}[htb]
\centering
\includegraphics[scale=0.40]{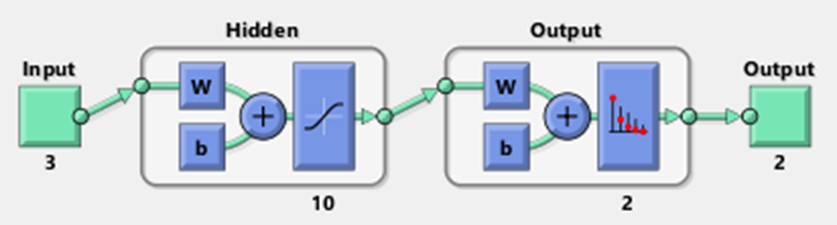}
\caption{Neural network used to learn the socialization level.}
\label{fig:RN}
\end{figure}

To compute the socialization level $\vartheta$ we use a classical supervised learning algorithm proposed by Moller \cite{Moller93}. The artificial neural network (ANN) (illustrated in Figure \ref{fig:RN}) uses a Scaled Conjugate Gradient (SCG) algorithm in the training process to calculate the socialization $\vartheta_i$ level for each individual $i$.

As described in Figure \ref{fig:RN}, the ANN has 3 inputs (collectivity $\phi_i$ of person $i$, mean Euclidean distance from a person $i$ to others $\bar{d_{i,j}}$ and the number of people in the Social Space\footnote{Social space is related to $3.6$ meters~\cite{hall98}.} according to Hall's proxemics~\cite{hall98} around the person $n_i$). In addition, the network has 10 hidden layers and 2 outputs (the probability of socialization and the probability of non socialization). The final accuracy from the training processes was 96\%. We used 16.000 samples (70\% of training and 30\% of validating). These samples %SO: todos?
were obtained from the 25 initial frames from each of the videos from our dataset. The remaining frames were used to test the ANN as described in Section~\ref{sec:results}.
%SO: não teve oersonal inspection

The ground truth (GT) was generated as follows: Firstly, we define if a person has a high socialization level $GT\_\vartheta_i$ based on Hall's proxemics, calculated according to the Equation~\ref{eq:socialization}:

\begin{equation}
GT\_\vartheta_i = \left\{
\begin{array}{ll}
0, & \text{~if~} n_i = 0 \\
\frac{n_i}{\rho}, & \text{~otherwise}
\end{array},
\label{eq:socialization}
\right .
\end{equation}

where $n_i$ is the number of individuals in the social space around the person $i$ and $\rho$ is the number of individuals in the analyzed frame. If $GT\_\vartheta_i >= 0.5$, we considered this person as a ``social'' person, otherwise the person is considered ``not social'' in the training processes. Secondly, we proceed a visual inspection manually correcting false positives or false negatives in comparison to our personal opinion. Using this GT and the neural network we evaluate  $\vartheta_i$ for each individual $i$ at each frame, for each video in the test group.

Once we get the socialization level $\vartheta_i$, we compute the isolation level $\varphi_i = 1 - \vartheta_i$, that corresponds to its inverse.

%where $np_{social}$ is the number of individuals in the social space, $d$ is a simple function to calculate the Euclidean distance between two individuals $i$ and $j$ ($x_i$ and $x_j$ are, respectively, the positions of individuals $i$ and $j$), $d_{Hall} = 3.6$ is the distance around an individual that represents its personal social space and $n$ is number of people in social space. The socialization level $\vartheta_i$ of an individual $i$ is calculated according to the Equation~\ref{eq:socialization}:

Finally, for each individual $i$ in a frame $f$ of a certain video $v$, we will have a features vector $\vec{V_i^{f,v}} = \left [x_i^{f,v}, s_i^{f,v}, \alpha_i^{f,v},  \phi_i^{f,v}, \vartheta_i^{f,v},  \varphi_i^{f,v} \right ]$. Then, computing the average for individual $i$, for all frames of a video $v$, we will have vector $\vec{V_i^v}$ for each person $i$.

%, and computing the average for all individuals of a certain video we have a feature vector $\vec{V^v}$.
In this paper, we are interested about mapping the features vector from each individual in a specific video $\vec{V_i^v}$ to OCEAN dimensions, detailed in next section. 

% computed the average for all frames and generate a vector $\vec{V_i}$ of extracted data where $\vec{V_i} = \left [x_i, s_i, \alpha_i,  \phi_i, \vartheta_i,  \varphi_i \right ]$. In the next section we describe how these features are mapped into OCEAN dimensions.

% For groups detection, we compute the following parameters for each pair of agents $i$ and $j$: $s(v_i,v_j)$, $o(\alpha_i,\alpha_j)$ and $d(x_i,x_j)$, where $s(v_i,v_j)$, $o(\alpha_i,\alpha_j)$ are the differences of speed and orientation and $d(x_i,x_j)$ is the Euclidean distance between the two individuals. We use the notion of distances based on the "\textit{proxemics}" described  by Hall~\cite{Hall:1990} to define that two agents belong to the same group according with three tests: If $(d(x_i,x_j) <= 1.2 \text{meter})$ and $(o(\alpha_i,\alpha_j) <= 15^{\circ})$ and ($s(v_i,v_j) < \beta$), where $\beta=5\%$ was empirically defined. Based on this rule, agents are grouped in pairs. In the next step, we check which pairs have one individual in common, and merge them into larger groups. This process is performed until the group formation does not share individuals, i.e. they are disjoint.

\subsection{Mapping crowd features in Cultural Dimensions}

Our goal is to map data from  $\vec{V_i}$ to $\vec{BF_i}$, where the last one is related to the Big-Fve dimensions (or OCEAN) for each individual $i$ for a certain video and described as a features vector: $\vec{BF_i} = \left [O_i, C_i, E_i, A_i, N_i \right ]$.

Therefore, in our method $\vec{BF}$ is computed based on NEO PI-R. With human beings, OCEAN is calculated based on their answers to the full version of NEO PI-R, with $240$ items. Our goal is to find out NEO PI-R ``answers'' for each individual in the video sequence, based on their features ($\vec{V_i}$). So, we have proposed a series of empirically defined equations to map individual and crowd characteristics (in video sequences) to OCEAN cultural dimensions.

As stated before, the complete version of NEO PI-R has $240$ items. Firstly, we selected 25 items from NEO PI-R inventory that had a direct relationship with crowd behavior.  From the 25 items selected, 18 (72\%) are from \textit{Extroversion}, 3 (12\%) are from \textit{Neuroticism}, 2 (8\%) are from \textit{Agreeableness}, 1 (4\%) is from \textit{Openness} and 1 (4\%) is from \textit{Conscientiousness}. One example of items presented in NEO PI-R is ``1 - Have clear goals, work to them in orderly way'' and possible answers are in the interval [0;4] which respectively represent: Strongly Disagree, Disagree, Neutral, Agree and Strongly Agree.

Our proposal is to answer these 25 items (see Table~\ref{tab:equations}) for each individual at each frame in the video through the Equations on the right in Table~\ref{tab:equations}. For example, in order to represent the item ``1 - Have clear goals, work to them in orderly way'', we consider that the individual $i$ should have a high velocity $s_i$ and low angular variation $\alpha_i$ to have answer compatible with 4. So the equation for this item is $Q_1 = s_i+\frac{1}{\alpha_i}$. In this way, we empirically defined equations for all 25 items, as presented in Table~\ref{tab:equations}.

\begin{table}[h]
   \renewcommand{\arraystretch}{1.45}
   \centering
   \scriptsize
   \caption{Equations from each NEO PI-R item selected.}
   \begin{adjustbox}{max width=\textwidth}
     \begin{tabular}{l|c}
        \hline\noalign{\smallskip}
        NEO PI-R Item & Equation \\
        \noalign{\smallskip}
        \hline
        \noalign{\smallskip}
        \hline
        \noalign{\smallskip}
        1 - Have clear goals, work to them in orderly way & $Q_1 = s_i+\frac{1}{\alpha_i}$ \\
         \noalign{\smallskip}
        \hline
        \noalign{\smallskip}
        2. Follow same route when go somewhere & $Q_2 = \alpha_i$ \\
        \noalign{\smallskip}
        \hline
        \noalign{\smallskip}
        3. Shy away from crowds & \multirow{6}{*}{$Q_{3-8} = \varphi_i$} \\
        4. Don’t get much pleasure chatting with people & \\
        5. Usually prefer to do things alone & \\
        6. Prefer jobs that let me work alone, unbothered & \\
        7. Wouldn’t enjoy holiday in Las Vegas & \\
        8. Many think of me as somewhat cold, distant & \\
        \noalign{\smallskip}
        \hline
        \noalign{\smallskip}
        9. Rather cooperate with others than compete & \multirow{2}{*}{$Q_{9-10}=\phi_i$} \\
        10. Try to be courteous to everyone I meet & \\
        \noalign{\smallskip}
        \hline
        \noalign{\smallskip}
        11. Social gatherings usually bore me & $Q_{11}=\varphi_i + std(\alpha_i)$ \\
        \noalign{\smallskip}
        \hline
        \noalign{\smallskip}
        12. Usually seem in hurry & $Q_{12}=s_i+\alpha_i$  \\
        \noalign{\smallskip}
        \hline
        \noalign{\smallskip}
        13. Often disgusted with people I have to deal with & $Q_{13}=\varphi_i+\frac{1}{\phi_i}$ \\
        \noalign{\smallskip}
        \hline
        \noalign{\smallskip}
        14. Have often been leader of groups belonged to & $Q_{14}=\phi_i+\vartheta_i+\frac{1}{\alpha_i}$  \\
        \noalign{\smallskip}
        \hline
        \noalign{\smallskip}
        15. Would rather go my own way than be a leader & $Q_{15}=\frac{1}{Q_{14}}$ \\
        \noalign{\smallskip}
        \hline
        \noalign{\smallskip}
        16. Like to have lots of people around me & \multirow{6}{*}{$Q_{16-21}=\vartheta_i$} \\
        17. Enjoy parties with lots of people & \\
        18. Like being part of crowd at sporting events & \\
        19. Would rather a popular beach than isolated cabin & \\
        20. Really enjoy talking to people & \\
        21. Like to be where action is & \\
        \noalign{\smallskip}
        \hline
        \noalign{\smallskip}
        22. Feel need for other people if by myself for long & \multirow{4}{*}{$Q_{22-25}=\vartheta_i+\phi_i$} \\
        23. Find it easy to smile, be outgoing with strangers & \\
        24. Rarely feel lonely or blue & \\
        25. Seldom feel self-conscious around people & \\
        \noalign{\smallskip}
        \hline
     \end{tabular}
   \end{adjustbox}
   \label{tab:equations}
\end{table}

Once all questions $k$ (in the interval $[1;25]$) have been answered for all individuals $i$, we will have $\vec{Q_{i,k}^f}$ for each frame $f$. We computed the average values to have one vector $\vec{Q_{i,k}}$ per video.

As already mentioned, NEO PI-R items answers vary from $0$ to $4$. We converted the values obtained in $\vec{Q_{i,k}}$ in one of the 5 score possible options (0, 1, 2, 3 and 4) by simply normalizing the answers in 5 uniformly distributed levels, since we know the maximum level for each item at each video. We called this normalized vector as $\vec{Q'_{i,k}}$. In NEO PI-R definitions, some questions should invert the values, because an item score 4 (Strongly Agree) can represent a high value of Extroversion or low, depending on the question. For example, let us  analyze questions 4 and 16. A score=4 to both of them represents completely opposite answers in terms of sociability. So, to get the correct values, we applied a factor to the questions which score should be inverted: $\vec{Q^*_{i,k}} = 4 - \vec{Q'_{i,k}}$.

In addition, in NEO PI-R definition, each of the questions $\vec{Q'_{k}}$ are associated to one of the Big Five dimensions, as shown in next equations:

\begin{equation}
O_i = \frac{Q^*_{i,2}}{\varrho},
\end{equation}
\begin{equation}
C_i = \frac{Q'_{i,1}}{\varrho},
\end{equation}
\begin{equation}
E'_i = Q'_{i,3} + Q'_{i,12} + Q'_{i,14} + \sum_{q=16}^{23} Q'_{i,q},
\label{eq:e1}
\end{equation}
\begin{equation}
E^*_i = \sum_{q=4}^{8} Q^*_{i,q} + Q^*_{i,11} +  Q^*_{i,15},
\label{eq:e2}
\end{equation}
\begin{equation}
E_i = \frac{(E'_i+ E^*_i)}{\varrho},
\label{eq:e3}
\end{equation}
\begin{equation}
A_i = \frac{\sum_{q=9}^{10}Q'_{i,q}}{\varrho},
\end{equation}
\begin{equation}
N_i = \frac{Q'_{i,13}+ \sum_{q=24}^{25}Q^*_{i,q}}{\varrho},
\end{equation}

where $\varrho$ represents the percentage of questions from the total, in each dimension (O, C, E, A and N), respectively 4\%, 4\%, 72\%, 8\% and 12\%, as explained previously.

Once we get the OCEAN values of each person, we calculate the OCEAN of the video by the mean of people's OCEAN. In a similar way, the OCEAN of a country is the mean of videos from that country. In the next section we present some obtained results of our method.

\section{Experimental Results}
\label{sec:results}

\begin{figure}[h]
\center
\subfigure[fig:H_E][Higher E (Brazil)]{\includegraphics[scale=0.235]{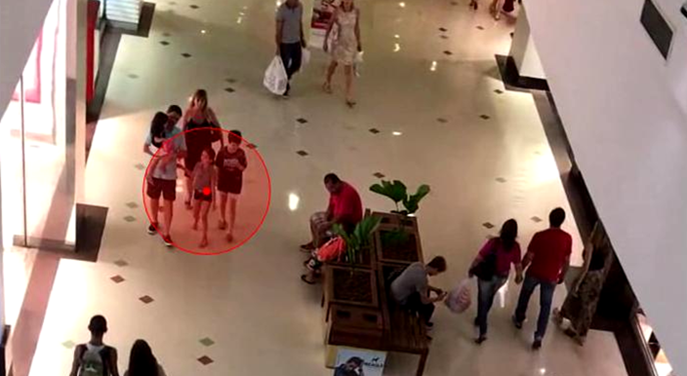}}
\subfigure[fig:L_E][Lower E - Higher N (China)]{\includegraphics[scale=0.235]{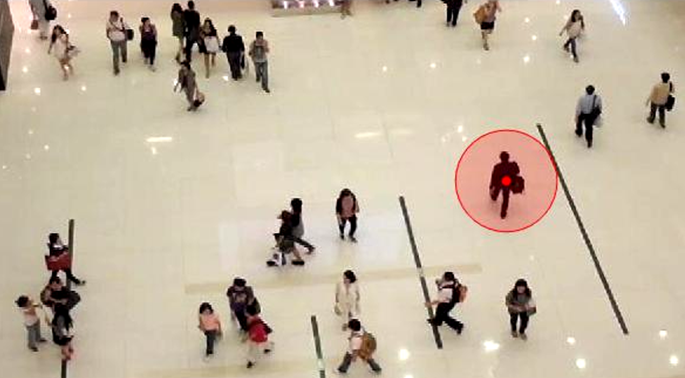}}
\qquad
\subfigure[fig:H_A][Higher A (Brazil)]{\includegraphics[scale=0.235]{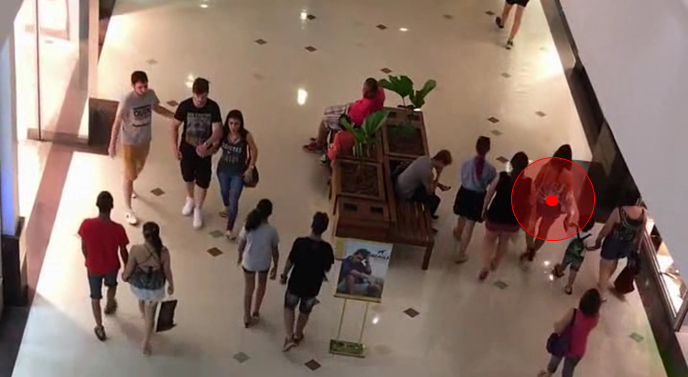}}
\subfigure[fig:L_A][Lower A (China)]{\includegraphics[scale=0.235]{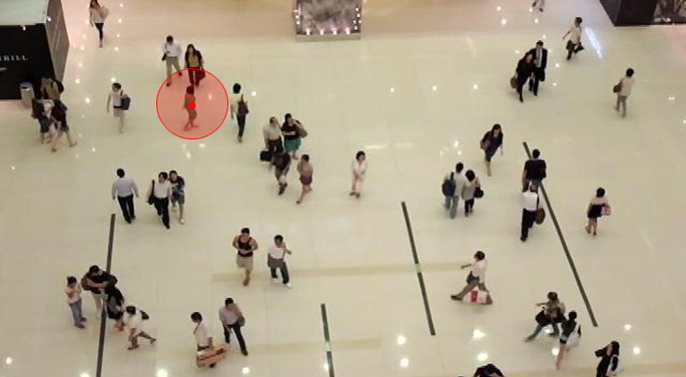}}
\qquad
\subfigure[fig:H_O][Higher O (Japan)]{\includegraphics[scale=0.235]{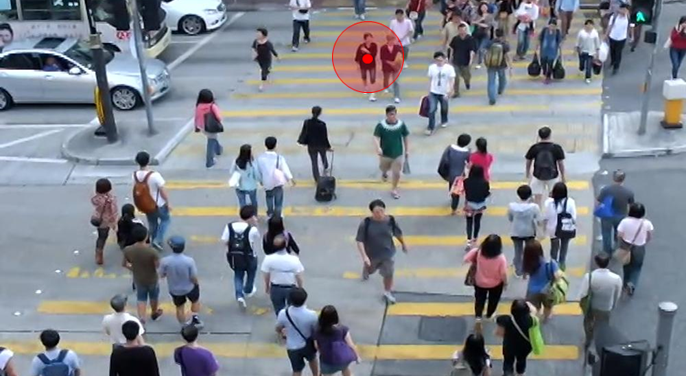}}
\subfigure[fig:L_O][Lower O (Brazil)]{\includegraphics[scale=0.235]{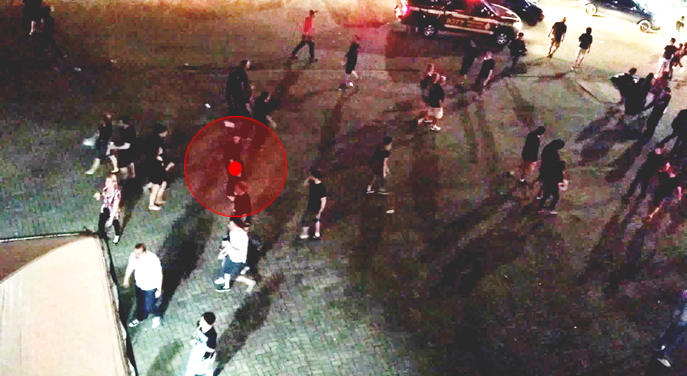}}
\caption{Examples of some individuals OCEAN levels: a) the highlight person has the highest Extraversion, b) shows the person with the lowest Extraversion (and highest Neuroticism), c) shows the person with the highest Agreeableness and the person highlighted in d) has the lowest Agreeableness. The highlight person in e) has the highest Openness and the person highlighted in d) has the lowest Openness.}
\label{fig:hig_low_E_A}
\end{figure}

In this section we discuss some results obtained with our approach. We evaluated our method in a set of $20$ videos from $4$ countries ($9$ from Brazil, $5$ from China, $3$ from Austria and $3$ from Japan). These videos, with a duration varying between 100 and 900 frames, were collected from different public databases available on the Internet, such as \cite{Zhou2014, Caviar:2016, Rodriguez:2011}. Firstly, we get the OCEAN of each individual in the scene (Figure \ref{fig:hig_low_E_A} shows some examples). In Figure~\ref{fig:hig_low_E_A} (a) we can observe the higher E that was found in an individual, part of a group of people, while the opposite happens in (b) when lower E was computed for individual alone and far from the others. 

\begin{figure*}[htb]
\centering
\includegraphics[scale=0.53]{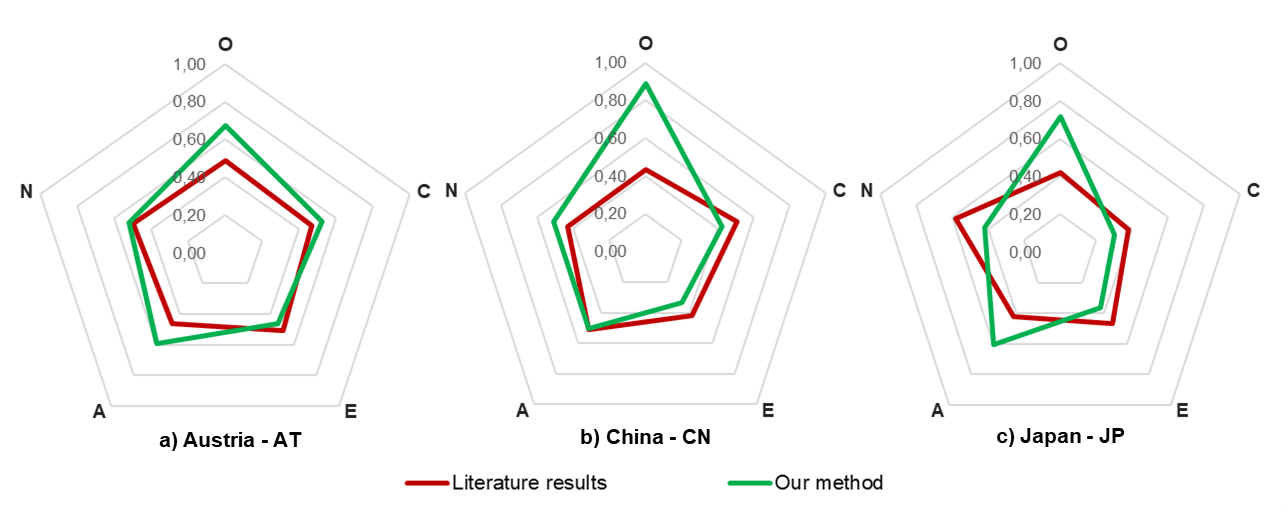}
%\vspace{-0.5cm}
\caption{OCEAN comparison between our approach and literature values.}
\label{fig:all_countries}
\end{figure*}

Same kind of analysis can be done for images (c) and (d) relating to their collectivity (higher and lower respectively) as described in Equation~\ref{eq:colec}. Although it is more difficult to visual inspect the dimensions O, C and N we present the qualitative results. For example in Figure~\ref{fig:hig_low_E_A} (e) the highlighted individual has lower angular variation in comparison to all others (higher O value), while in (f) this is the individual with higher angular variation, consequently having lower value of O. In addition, in Figure~\ref{fig:hig_low_E_A} (b) we obtained the higher value of N, since it is dependent of the inverse of collectivity and socialization. Once the individual OCEAN values are computed, we get the mean OCEAN value for each video. The country's OCEAN, in turn, is calculated by the average OCEANs of that country's videos.

\begin{figure}[htb]
\centering
\includegraphics[scale=0.4]{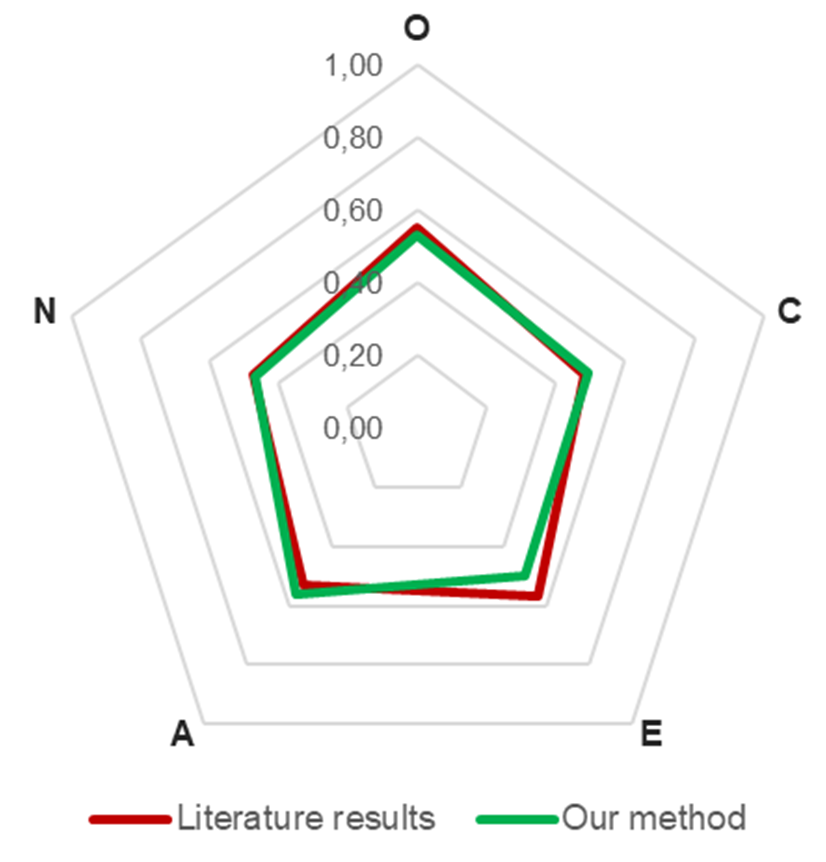}
\caption{OCEAN results from Brazil.}
\label{fig:OCEAN_BR}
\end{figure}

Figure~\ref{fig:OCEAN_BR} shows the results obtained by the country \textit{Brazil} in all OCEAN dimensions, in comparison with the literature~\cite{costa07}, considered as ground-truth in our approach. It is interesting to highlight that results achieved for this country showed the higher accuracy, when compared to the other countries (see the Figure \ref{fig:all_countries}). This was the country with more available videos to be processed in our method (9 videos), in comparison with other countries.

\begin{figure}[htb]
\centering
\includegraphics[scale=0.45]{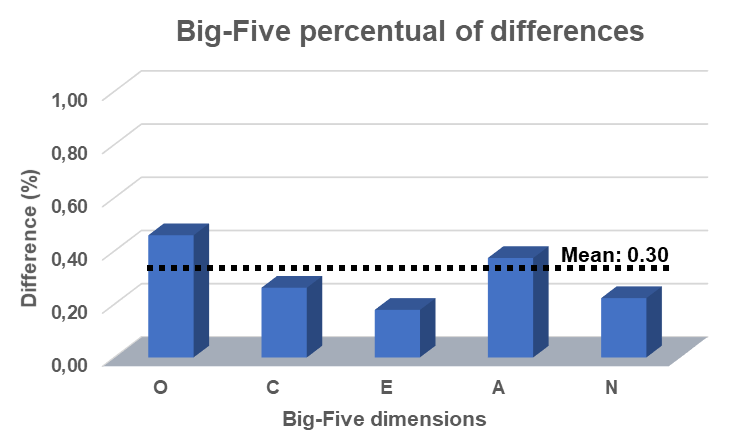}
\caption{Big-Five percentual of differences.}
\label{fig:differences_ocean}
\end{figure}

In addition, we computed the perceptual error when accumulating each dimension from all videos and compared with literature for those Countries. Figure~\ref{fig:differences_ocean} shows such errors and also indicates that the presented error of dimension \textit{E} has lower value; that is an interesting observation since this was the dimension that had more questions to be analyzed, as shown in Equations~\ref{eq:e1}, ~\ref{eq:e2} and ~\ref{eq:e3}.

In terms of cultural aspects of individuals in the videos, Table~\ref{tab:high_low} shows the countries that get the higher and lower values in each dimension, according to our approach. For example, Brazil is the most extrovert country, while the less neurotic is Japan. 

\begin{table}[h]
   \renewcommand{\arraystretch}{1.5}
   \centering \small
   \caption{Countries with Higher and lower values in each Big Five dimension.}
   \begin{adjustbox}{max width=\textwidth}
     \begin{tabular}{c|c|c|c|c|c}
        \hline\noalign{\smallskip}
         & \textbf{O} & \textbf{C} & \textbf{E} & \textbf{A} & \textbf{N} \\
        \noalign{\smallskip}
        \hline
        \noalign{\smallskip}
        \hline
        \noalign{\smallskip}
        \multirow{2}{*}{\textbf{Higher}} & China & Austria & Brazil & Japan & Austria \\
        & $(0.89)$ & $(0.53)$ & $(0.50)$ & $(0.60)$ & $(0.52)$ \\
        \noalign{\smallskip}
        \hline
        \noalign{\smallskip}
        \multirow{2}{*}{\textbf{Lower}} & Brazil & Japan & China & China & Japan \\
        & $(0.53)$ & $(0.30)$ & $(0.33)$ & $(0.51)$ & $(0.42)$ \\
        \noalign{\smallskip}
        \hline
     \end{tabular}
   \end{adjustbox}
   \label{tab:high_low}
\end{table}

\begin{figure}[htb]
\centering
\includegraphics[scale=0.39]{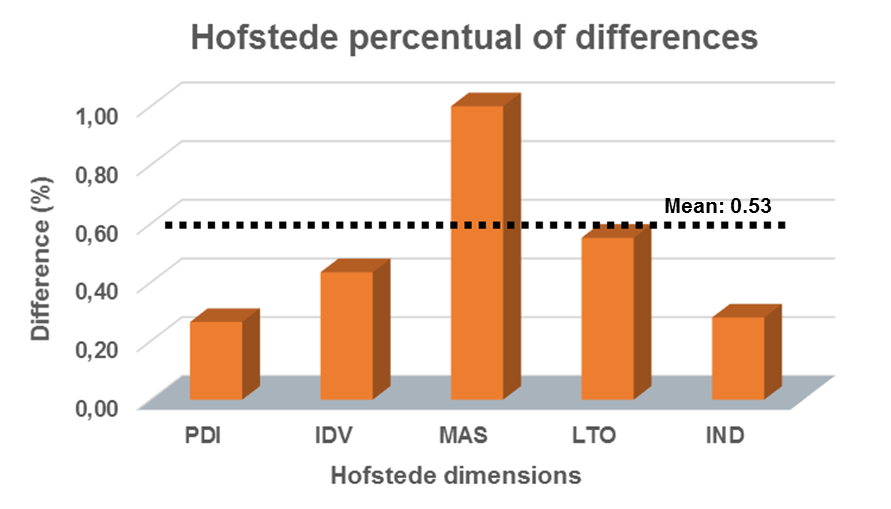}
\caption{Hofstede percentual of differences.}
\label{fig:differences_hofs}
\end{figure}

According to previous work, another classical cultural dimension is proposed by Hofstede~\cite{Hofstede:2011}. In a recent paper, Favaretto et al. ~\cite{Favaretto:2016} presented the cultural aspects of people in video using Hofstede's cultural dimensions theory (Figure \ref{fig:differences_hofs}). We compared our error using Big-Five (Figure~\ref{fig:differences_ocean}) with this method, when using Hofstede's.

The accuracy of each approach (OCEAN and Hofstede) can be found in terms of the mean difference percentual when compared with the literature results, considering all dimensions among all videos. With an average error of 30\% from the results presented in literature, the OCEAN method proved to be more promising than Hofstede (with an average error of 53\%) for culturality mapping.

% Comparing with previous work with similar goals, we analysed the perceptual error of each OCEAN dimension in this paper (Figure \ref{fig:differences_ocean}) with the results obtained with Hofstede analysis~\cite{Favaretto:2016}, as showed in Figure \ref{fig:differences_hofs}.

It is important to note that the mapping to OCEAN dimensions was empirically defined through equations using data extracted from computer vision. NEO PI-R measured these dimensions by considering a different type of information (subjective responses of individuals collected through questionnaires).

In this sense, it is possible to affirm that, even with few videos used, the results obtained with the proposed approach are coherent with NEO PI-R results and more effective if compared with Hofstede dimensions. The factor Extroversion (E) is the one that seems to be more predictable with our model. Probably because this factor comprehends the majority of items related to crowd behaviors. 

\section{Conclusions}
\label{sec:conclusions}

In this paper we described a way to map equations to compute individual-level traits from video sequences, based on individuals and groups features. Our model computed, from video sequences, OCEAN personality traits and compared with data from different countries existent in the literature. In addition, we compared with some previous work that computed Hofstede dimensions using a similar approach. We believe the results are promising and video sequences can be used to detect crowd cultural aspects.

For our future work we intend to validate our model asking participants how much they agree with the assigned score of each item in the Big-Five questionnaire that resulted in our model for individuals with high scores in selected videos. By doing this we can compare human score with computer generated score of the same videos.

We also intend to increase our set of video data. Both aspects, number of countries and the among of videos from each of them, should be considered. One of the major difficulties of this work was to find a suitable set of videos to perform the experiments.

In addition, we intend to make video-recordings of group situations where each individual presented in the video has previously evaluated OCEAN scores. For this, one plausible option is evaluate our method with the SALSA dataset \cite{Salsa:2016}, which provides Big-five personality traits for a group of people in video sequences. 

We may thus have another evidence of the validity of the presented model. In addition, we plan to create a new model comprehending different psychological domains related to crowd characteristics that have documented cultural differences: extraversion from the big-five model \cite{costa07}, Hofstede’s collectivism \cite{Hofstede:2011}, Hall´s personal space \cite{Hall:1990}, fundamental diagram \cite{CHATTARAJ2009} and the subjective pace of time \cite{Levine:1999}. 

\bibliographystyle{IEEEtran}
\bibliography{refs_bib}

\begin{thebibliography}{10}
\providecommand{\url}[1]{#1}
\csname url@samestyle\endcsname
\providecommand{\newblock}{\relax}
\providecommand{\bibinfo}[2]{#2}
\providecommand{\BIBentrySTDinterwordspacing}{\spaceskip=0pt\relax}
\providecommand{\BIBentryALTinterwordstretchfactor}{4}
\providecommand{\BIBentryALTinterwordspacing}{\spaceskip=\fontdimen2\font plus
\BIBentryALTinterwordstretchfactor\fontdimen3\font minus
  \fontdimen4\font\relax}
\providecommand{\BIBforeignlanguage}[2]{{%
\expandafter\ifx\csname l@#1\endcsname\relax
\typeout{** WARNING: IEEEtran.bst: No hyphenation pattern has been}%
\typeout{** loaded for the language `#1'. Using the pattern for}%
\typeout{** the default language instead.}%
\else
\language=\csname l@#1\endcsname
\fi
#2}}
\providecommand{\BIBdecl}{\relax}
\BIBdecl

\bibitem{Chan2009}
A.~B. Chan and N.~Vasconcelos, ``Bayesian poisson regression for crowd
  counting,'' in \emph{12th IEEE ICCV}, Sept 2009, pp. 545--551.

\bibitem{cai2014}
Z.~Cai, Z.~L. Yu, H.~Liu, and K.~Zhang, ``Counting people in crowded scenes by
  video analyzing,'' in \emph{9th IEEE ICIEA}, June 2014, pp. 1841--1845.

\bibitem{Solmaz2012}
\BIBentryALTinterwordspacing
B.~Solmaz, B.~E. Moore, and M.~Shah, ``Identifying behaviors in crowd scenes
  using stability analysis for dynamical systems,'' \emph{IEEE PAMI}, vol.~34,
  no.~10, pp. 2064--2070, Oct. 2012. [Online]. Available:
  \url{http://dx.doi.org/10.1109/TPAMI.2012.123}
\BIBentrySTDinterwordspacing

\bibitem{Zhou2014}
B.~Zhou, X.~Tang, H.~Zhang, and X.~Wang, ``Measuring crowd collectiveness,''
  \emph{IEEE PAMI}, vol.~36, no.~8, pp. 1586--1599, Aug 2014.

\bibitem{Ricky:15}
\BIBentryALTinterwordspacing
R.~J. Sethi, ``Towards defining groups and crowds in video using the atomic
  group actions dataset,'' in \emph{2015 {IEEE} International Conference on
  Image Processing, {ICIP} 2015, Quebec City, QC, Canada, September 27-30,
  2015}, 2015, pp. 2925--2929. [Online]. Available:
  \url{http://dx.doi.org/10.1109/ICIP.2015.7351338}
\BIBentrySTDinterwordspacing

\bibitem{jo2013review}
H.~Jo, K.~Chug, and R.~J. Sethi, ``A review of physics-based methods for group
  and crowd analysis in computer vision,'' \emph{Journal of Postdoctoral
  Research}, vol.~1, no.~1, pp. 4--7, 2013.

\bibitem{solera_2013}
F.~Solera, S.~Calderara, and R.~Cucchiara, ``Structured learning for detection
  of social groups in crowd,'' in \emph{10th IEEE AVSS}, Aug. 2013.

\bibitem{Shao2014}
J.~Shao, C.~C. Loy, and X.~Wang, ``Scene-independent group profiling in
  crowd,'' in \emph{IEEE CVPR}, June 2014, pp. 2227--2234.

\bibitem{Feng2015}
L.~Feng and B.~Bhanu, ``Understanding dynamic social grouping behaviors of
  pedestrians,'' \emph{IEEE STSP}, vol.~9, no.~2, pp. 317--329, March 2015.

\bibitem{Chandran2015}
A.~Chandran, L.~A. Poh, and P.~Vadakkepat, ``Identifying social groups in
  pedestrian crowd videos,'' in \emph{ICAPR}, Jan 2015, pp. 1--6.

\bibitem{CHATTARAJ2009}
\BIBentryALTinterwordspacing
U.~Chattaraj, A.~Seyfried, and P.~Chakroborty, ``Comparison of pedestrian
  fundamental diagram across cultures,'' \emph{ACS}, vol.~12, no.~03, pp.
  393--405, 2009. [Online]. Available:
  \url{http://www.worldscientific.com/doi/abs/10.1142/S0219525909002209}
\BIBentrySTDinterwordspacing

\bibitem{Favaretto:2016}
R.~Favaretto, L.~Dihl, R.~Barreto, and S.~R. Musse, ``Using group behaviors to
  detect hofstede cultural dimensions,'' in \emph{IEEE International Conference
  on Image Processing (ICIP)}, 2016.

\bibitem{Hofstede:2011}
G.Hofstede, ``Dimensionalizing cultures: The hofstede model in context,''
  \emph{ScholarWorks@GVSU. Online Readings in Psychology and Culture}.

\bibitem{costa07}
P.~T.~C. Jr and R.~R. McCrae, \emph{NEO PI-R: inventário de personalidade NEO
  revisado.}, .Vetor Editora, 2007, são Paulo.

\bibitem{costa92}
P.~T.~C. Jr. and R.~R. McCrae, \emph{Revised NEO Personality Inventory
  (NEO-PI-R) and the NEO Five-Factor Inventory (NEO-FFI) professional manual.},
  Odessa, FL: Psychological Assessment Resources., 1992.

\bibitem{cattell50}
R.~B. Cattell, \emph{Personality: A systematic, theoretical, and factual
  study}, 1st~ed.\hskip 1em plus 0.5em minus 0.4em\relax New York: McGraw-Hill,
  1950.

\bibitem{hall98}
G.~L. C.~S.~Hall and J.~B. Campbell, \emph{Theories Of Personality},
  4th~ed.\hskip 1em plus 0.5em minus 0.4em\relax New Jersey: John Wiley \&
  Sons, 1998.

\bibitem{digman90}
J.~M. Digman, ``Personality structure: Emergence of the five-factor model,''
  \emph{Annual Review of Psychology}, vol.~41, no.~1, pp. 417--440, 1990.

\bibitem{lordw07}
W.~Lord, \emph{Neo Pi-R – A Guide to Interpretation and Feedback in a Work
  Context}, 1st~ed.\hskip 1em plus 0.5em minus 0.4em\relax Hogrefe Ltd, 2007.

\bibitem{cattell48}
\BIBentryALTinterwordspacing
R.~B. Cattell, ``The primary personality factors in women compared with those
  in men,'' \emph{British Journal of Statistical Psychology}, vol.~1, no.~2,
  pp. 114--130, 1948. [Online]. Available:
  \url{http://dx.doi.org/10.1111/j.2044-8317.1948.tb00231.x}
\BIBentrySTDinterwordspacing

\bibitem{fiske48}
\BIBentryALTinterwordspacing
D.~W. Fiske, \emph{Consistency of The Factorial Structures in Personality
  Ratings from Different Sources.}\hskip 1em plus 0.5em minus 0.4em\relax
  University of MICHIGAN, 1948. [Online]. Available:
  \url{https://books.google.com.br/books?id=o74lnQEACAAJ}
\BIBentrySTDinterwordspacing

\bibitem{McCrae2002}
R.~R. McCrae, \emph{NEO-PI-R Data from 36 Cultures}.\hskip 1em plus 0.5em minus
  0.4em\relax Boston, MA: Springer US, 2002, pp. 105--125.

\bibitem{McCrae05}
\BIBentryALTinterwordspacing
R.~McCrae and A.~Terracciano, ``Universal features of personality traits from
  the observer's perspective : data from 50 cultures,'' \emph{Journal of
  Personality and Social Psychology}, vol.~88, no.~3, pp. 547--561, March 2005.
  [Online]. Available: \url{http://sro.sussex.ac.uk/14937/}
\BIBentrySTDinterwordspacing

\bibitem{Barry97}
B.~Barry and G.~L. Stewart, ``Composition, process, and performance in
  self-managed groups: The role of personality,'' \emph{Journal of Applied
  Psychology}, vol.~82, p.~62, feb 1997.

\bibitem{kaup06}
D.~Kaup, T.~Clarke, L.~Malone, and R.~Oleson, ``Society for computer
  simulation,'' \emph{SIMULATION SERIES}, vol.~38, no.~4, pp. 365--370, 2006.

\bibitem{Durupinar08}
\BIBentryALTinterwordspacing
F.~Durupinar, J.~Allbeck, N.~Pelechano, and N.~Badler, ``Creating crowd
  variation with the ocean personality model,'' in \emph{Proc. of the 7th
  International Joint Conf. on Autonomous Agents and Multiagent Systems -
  Volume 3}.\hskip 1em plus 0.5em minus 0.4em\relax Richland, SC: IFAAMAS,
  2008, pp. 1217--1220. [Online]. Available:
  \url{http://dl.acm.org/citation.cfm?id=1402821.1402835}
\BIBentrySTDinterwordspacing

\bibitem{Guy11}
\BIBentryALTinterwordspacing
S.~J. Guy, S.~Kim, M.~C. Lin, and D.~Manocha, ``Simulating heterogeneous crowd
  behaviors using personality trait theory,'' in \emph{Proceedings of the 2011
  ACM SIGGRAPH/Eurographics Symposium on Computer Animation}, ser. SCA
  '11.\hskip 1em plus 0.5em minus 0.4em\relax New York, USA: ACM, 2011, pp.
  43--52. [Online]. Available: \url{http://doi.acm.org/10.1145/2019406.2019413}
\BIBentrySTDinterwordspacing

\bibitem{Kosinski:2013}
M.~Kosinski, D.~Stillwell, and T.~Graepel, ``Private traits and attributes are
  predictable from digital records of human behavior,'' \emph{Proceedings of
  the National Academy of Sciences}, vol. 110, no.~15, pp. 5802--5805, 2013.

\bibitem{paulhus:1992}
D.~L. Paulhus and M.~N. Bruce, ``The effect of acquaintanceship on the validity
  of personality impressions: A longitudinal study.'' \emph{Journal of
  Personality and Social Psychology}, vol.~63, no.~5, p. 816, 1992.

\bibitem{oh:2011}
I.-S. Oh, G.~Wang, and M.~K. Mount, ``Validity of observer ratings of the
  five-factor model of personality traits: a meta-analysis.'' 2011.

\bibitem{JASP1355}
H.~Aarts, B.~Verplanken, and A.~van Knippenberg, ``Predicting behavior from
  actions in the past: Repeated decision making or a matter of habit?''
  \emph{Journal of Applied Social Psychology}, vol.~28, no.~15, pp. 1355--1374,
  1998.

\bibitem{Yee:2011}
\BIBentryALTinterwordspacing
N.~Yee, N.~Ducheneaut, L.~Nelson, and P.~Likarish, ``Introverted elves \&\#38;
  conscientious gnomes: The expression of personality in world of warcraft,''
  in \emph{Proceedings of the SIGCHI Conference on Human Factors in Computing
  Systems}, ser. CHI '11.\hskip 1em plus 0.5em minus 0.4em\relax New York, NY,
  USA: ACM, 2011, pp. 753--762. [Online]. Available:
  \url{http://doi.acm.org/10.1145/1978942.1979052}
\BIBentrySTDinterwordspacing

\bibitem{Yee:2011-2}
\BIBentryALTinterwordspacing
N.~Yee, H.~Harris, M.~Jabon, and J.~N. Bailenson, ``The expression of
  personality in virtual worlds,'' \emph{Social Psychological and Personality
  Science}, vol.~2, no.~1, pp. 5--12, 2011. [Online]. Available:
  \url{http://dx.doi.org/10.1177/1948550610379056}
\BIBentrySTDinterwordspacing

\bibitem{Lala2012}
D.~Lala, S.~Thovuttikul, and T.~Nishida, ``Towards a virtual environment for
  capturing behavior in cultural crowds,'' in \emph{6th ICDIM}, Sept 2011, pp.
  310--315.

\bibitem{Hofstede:1991}
\BIBentryALTinterwordspacing
G.Hofstede, \emph{Cultures and organizations: software of the mind.}\hskip 1em
  plus 0.5em minus 0.4em\relax London: McGraw-Hill., 1991. [Online]. Available:
  \url{http://books.google.com.br/books?id=zGYPwLj2dCoC}
\BIBentrySTDinterwordspacing

\bibitem{Kaminka:2011}
N.~Fridman, A.~Zilka, and G.~A. Kaminka, ``The impact of cultural differences
  on crowd dynamics in pedestrian and evacuation domains,'' Bar Ilan
  University, Computer Science Department, {MAVERICK} Group, available at
  http://www.cs.biu.ac.il/$^\sim$galk/Publications/, Tech. Rep. MAVERICK
  2011/01, 2011.

\bibitem{Bins2013}
\BIBentryALTinterwordspacing
J.~Bins, L.~L. Dihl, and C.~R. Jung, ``Target tracking using multiple patches
  and weighted vector median filters,'' \emph{MIV}, vol.~45, no.~3, pp.
  293--307, Mar. 2013. [Online]. Available:
  \url{http://dx.doi.org/10.1007/s10851-012-0354-y}
\BIBentrySTDinterwordspacing

\bibitem{Favaretto:Sib:2016}
R.~M. Favaretto, L.~Dihl, and S.~R. Musse, ``Detecting crowd features in video
  sequences,'' in \emph{Proceedings of Conference on Graphics, Patterns and
  Images (SIBGRAPI)}.\hskip 1em plus 0.5em minus 0.4em\relax IEEE Computer
  Society´s Conference Publishing Services, 2016.

\bibitem{Moller93}
``A scaled conjugate gradient algorithm for fast supervised learning,''
  \emph{Neural Networks}, vol.~6, no.~4, pp. 525 -- 533, 1993.

\bibitem{Caviar:2016}
R.~Fisher, \emph{CAVIAR: Context Aware Vision using Image-based Active
  Recognition}, 2016 (accessed May 13, 2016),
  \url{http://homepages.inf.ed.ac.uk/rbf/CAVIAR/}.

\bibitem{Rodriguez:2011}
M.~Rodriguez, J.~Sivic, I.~Laptev, and J.-Y. Audibert, ``Data-driven crowd
  analysis in videos,'' in \emph{Proceedings of the International Conference on
  Computer Vision (ICCV)}, 2011.

\bibitem{Salsa:2016}
X.~Alameda-Pineda, J.~Staiano, R.~Subramanian, L.~Batrinca, E.~Ricci, B.~Lepri,
  O.~Lanz, and N.~Sebe, ``Salsa: A novel dataset for multimodal group behavior
  analysis,'' \emph{IEEE Transactions on Pattern Analysis and Machine
  Intelligence}, vol.~38, no.~8, pp. 1707--1720, Aug 2016.

\bibitem{Hall:1990}
\BIBentryALTinterwordspacing
E.~Hall, \emph{The Hidden Dimension}, ser. A Doubleday anchor book.\hskip 1em
  plus 0.5em minus 0.4em\relax Anchor Books, 1990. [Online]. Available:
  \url{http://books.google.com.br/books?id=zGYPwLj2dCoC}
\BIBentrySTDinterwordspacing

\bibitem{Levine:1999}
R.~V. Levine and A.~Norenzayan, ``The pace of life in 31 countries,''
  \emph{Journal of Cross-Cultural Psychology}, vol.~30, no.~2, pp. 178--205,
  1999.

\end{thebibliography}
%\begin{thebibliography}{1}

%\end{thebibliography}

\end{document}